\documentclass[preprint,12pt]{elsarticle}

\usepackage{amssymb}
\usepackage{amsmath}
\usepackage{subcaption}
\usepackage{booktabs}
\usepackage{hyperref}
\usepackage{amsmath,amssymb,amsfonts}
\usepackage{algorithmic}
\usepackage{comment}
\usepackage{graphicx}
\usepackage[hang,flushmargin]{footmisc}
\usepackage{placeins}
\usepackage{subcaption}
\usepackage{textcomp}
\usepackage[table,xcdraw]{xcolor}
\usepackage{xcolor}
\usepackage{booktabs}
\usepackage{multirow}
\usepackage{siunitx}
\usepackage{titlesec}
\usepackage{amsfonts}
\usepackage{amssymb}
\usepackage{hyperref}
\usepackage{cleveref}

\begin{document}

\begin{frontmatter}


\title{Removing Averaging: Personalized Lip-Sync Driven Characters Based on Identity Adapter}




\author[1]{Yanyu Zhu}

\ead{zhu-yy24@mails.tsinghua.edu.cn}


\affiliation[1]{organization={Shenzhen International Graduate School, Tsinghua Universiy},
            city={Shenzhen},
            country={China}}

\author[1]{Lichen Bai}
\ead{blc22@mails.tsinghua.edu.cn}
\affiliation[2]{organization={Tencent},
            city={Shenzhen},
            country={China}}

\author[1]{Jintao Xu}
\ead{xjt22@mails.tsinghua.edu.cn}
            


\author[1]{Hai-Tao Zheng\corref{cor1}}

\ead{zheng.haitao@sz.tsinghua.edu.cn}

\cortext[cor1]{Corresponding author} 

\author[2]{Wei Zhao}
\ead{zachwei@tencent.com}

\author[2]{Hong-Gee Kim}
\ead{hgkim@snu.ac.kr}
\affiliation[3]{organization={Seoul National University},
            city={Seoul},
            country={Korea}}

\author[1]{Ruitong Liu}
\ead{liurita@sz.tsinghua.edu.cn}


\begin{abstract}
Recent advances in diffusion-based lip-syncing generative models have demonstrated their ability to produce highly synchronized talking face videos for visual dubbing. Although these models excel at lip synchronization, they often struggle to maintain fine-grained control over facial details in generated images. In this work, we identify ``lip averaging'' phenomenon where the model fails to preserve subtle facial details when dubbing out-of-distribution portrait videos. This issue arises because the commonly used U-Net backbone primarily integrates audio features into visual representations in the latent space via cross-attention mechanisms and multi-scale fusion, but it struggles to retain fine-grained lip details in the generated faces. To address this issue, we propose \textbf{UnAvgLip}, which extracts identity embeddings from reference videos to generate highly faithful facial sequences while maintaining accurate lip synchronization. Specifically, our method comprises two primary components: (1) an Identity Perceiver module that encodes facial embeddings to align with conditioned audio features; and (2) an ID-CrossAttn module that injects facial embeddings into the generation process, enhancing the capability of identity retention. Extensive experiments demonstrate that, at a modest training and inference cost, UnAvgLip effectively mitigates the ``averaging'' phenomenon in lip inpainting, significantly preserving unique facial characteristics while maintaining precise lip synchronization. Compared with the original approach, our method demonstrates significant improvements of 5\% on the identity consistency metric and 2\% on the SSIM metric across two benchmark datasets (HDTF and LRW). \footnote{The training code is available at \url{https://github.com/pigmeetsomebody/UnAvgLip}.}

\end{abstract}



\begin{keyword}


lip sync \sep video generation \sep talking face generation
\end{keyword}

\end{frontmatter}


\begin{figure}[thbp]
  \centering
    \includegraphics[width=\linewidth]{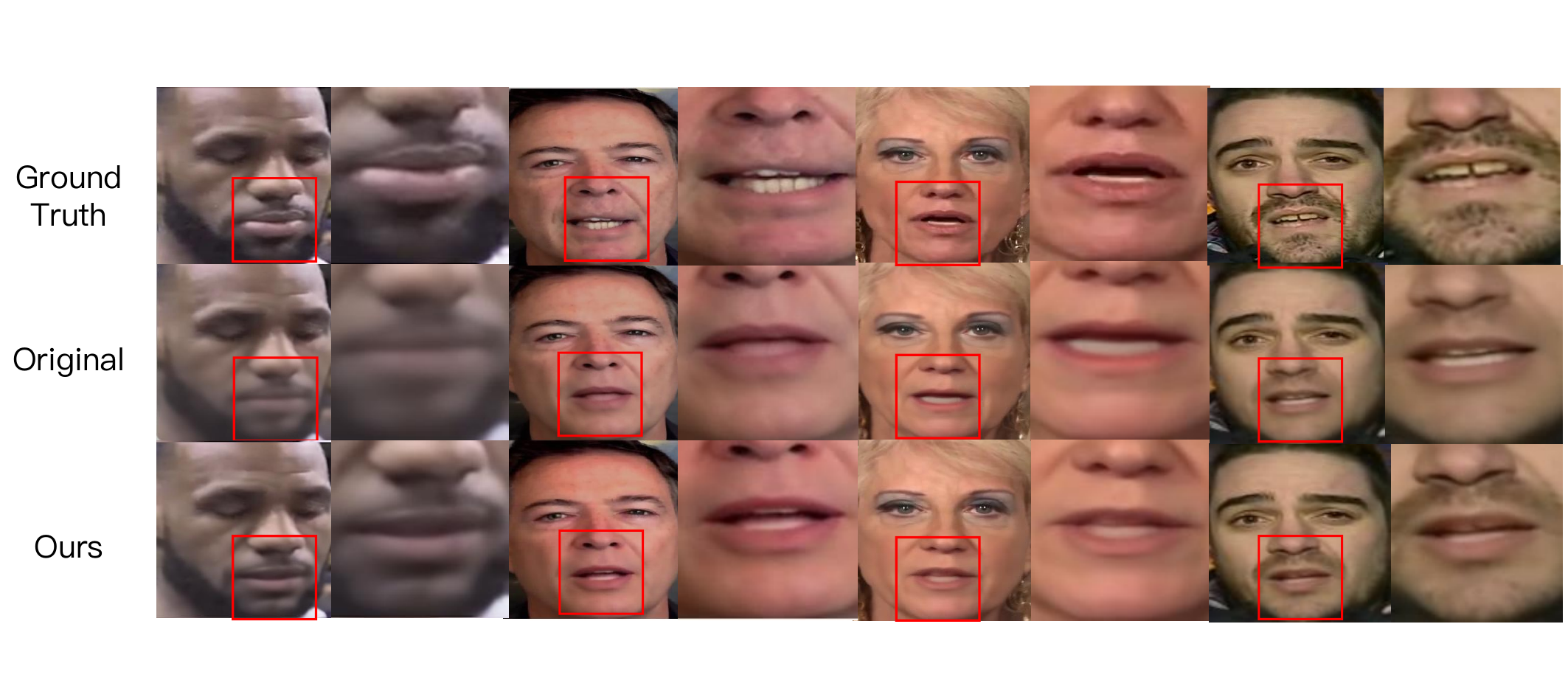}
  
  \caption{``Lip Averaging'' phenomenon. The original lip sync model fails to preserve finer details in lip region, resulting in average-looking lip, blurring teeth and mustache. By introducing identity attention into the original UNet model, our proposed UnAvgLip generates more detailed and refined facial features.}
  \label{fig:averaging_phenomenon}
\end{figure}

\section{Introduction}
Visual dubbing is the task of generating a talking face video where the lip movements and facial expressions of a speaker are synchronized with a given target audio. It aims to make the synthesized talking face appear as if the speaker is naturally pronouncing the provided speech, even if the original video was recorded in a different language or with different speech content. The progress of audio-driven talking face generation technology has received significant attention in recent years due to its wide range of applications, such  as human–computer intelligent interaction ~\citep{b30}, multi-language dubbing ~\citep{b27, bb34, b2}, and interactive avatars ~\citep{bb33, b28}.

Although recent advancements in generative networks, such as Generative Adversarial Networks (GANs)~\citep{bb4} and diffusion models~\citep{bb35, bb36}, have significantly advanced research in talking portrait video generation, visual dubbing still faces two major challenges: lip-speech synchronization and identity preservation. Lip-speech synchronization ensures that lip movements align accurately with phonemes  while identity preservation maintains the speaker’s facial appearance and expressions. Recent studies ~\citep{b24, b25, b5}  have leveraged the powerful generative capability of diffusion models to inpaint the lip region conditioned on audio. The new paradigm in diffusion-based visual dubbing involves masking the lip region in the source frame and then feeding it into the diffusion model to get inpainted results conditioned on encoded audio features. However, these methods often prioritize modality alignment between visual representation and phoneme over explicit identity retention, leading to blurry or generic lip shapes that deviate from the original speaker's identity. Additionally, performing lip-sync inpainting on the occluded regions of each frame independently may overlook certain global facial details such as lip shape, leading to a decline in identity consistency. We refer to this issue as the ``lip averaging'', where the model captures the general relationship between audio features and latent visual representations but fails to preserve finer details in the lip region, resulting in generic and homogenized lip reconstructions (as illustrated in ~\cref{fig:averaging_phenomenon}). Furthermore, customized talking portrait video generation with high ID fidelity often requires customized data collection followed by computationally intensive fine-tuning, which imposes significant constraints on scalability and real-world deployment.

\begin{figure}[htpb]
    \centering
    \includegraphics[width=\textwidth]{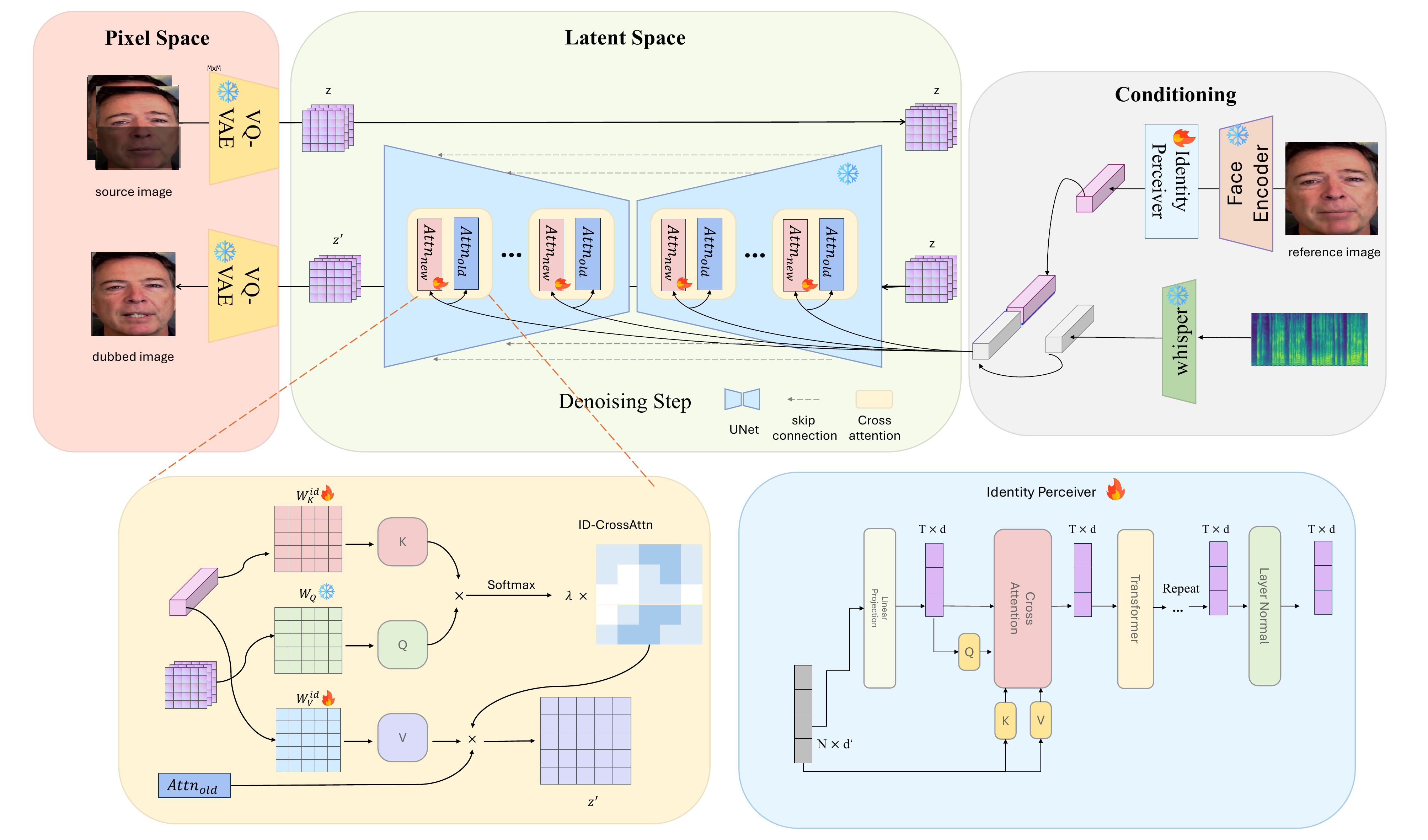}
    \caption{
    \textbf{Overview pipeline of our proposed UnAvgLip} that consists of two components: 1)The Identity Perceiver improves identity consistency in generated faces by projecting identity embeddings to the audio feature space with iterative attention. 2)The Adapter Module with decoupled cross-attention to inject encoded identity embedding into UNet as additional condition to improve the identity consistency of the generated face frames. We freeze all the parameters of UNet. Only Identity Perceiver and Adapter Module with total 48M parameters are trained. 
    }
    \label{fig:framewrok}
\end{figure}

To address these challenges, we propose \textbf{UnAvgLip}, a novel approach designed to enhance identity retention while maintaining precise lip-speech synchronization in diffusion-based visual dubbing. The comprehensive framework of the proposed UnAvgLip is illustrated in ~\cref{fig:framewrok}. Our framework explicitly incorporates identity information to ensure that generated talking face sequences preserve the speaker’s unique facial characteristics. By introducing a lightweight identity-adaptive mechanism, UnAvgLip effectively mitigates the trade-off between identity preservation and lip synchronization without requiring computationally expensive full-parameter fine-tuning. This design allows for a plug-and-play identity control module, enabling precise identity retention while ensuring natural and synchronized lip movements across frames. The contributions can be summarized as follows:

\begin{itemize}
\item We observe the ``lip averaging'' phenomenon in visual bubbling tasks, which arises due to inpainting lip region solely conditioned on audio, which inevitably weakens the capability of identity preservation.

\item We propose UnAvgLip, an efficient and powerful Lip-Sync framework that relies solely on a lightweight adapter, eliminating the need for full-parameter fine-tuning. This design enables a plug-and-play flexible module, effectively capitalizing on priors and ensuring precise lip synchronization while maintaining temporal consistency of facial identity information.

\item Rigorous experiments demonstrate the effectiveness of our method, achieving state-of-the-art performance in terms of visual quality and identity similarity i.e. over 2\% improvement in Structural Similarity (SSIM) and 5\% improvement in Identity Cosine Similarity (CSIM). Additionally, our approach performs comparably to or slightler better than the prior methods in terms of lip synchronization.

\end{itemize}

\section{Related Works}
\subsection{GAN-based Talking Face Generation.}
Generative adversarial networks (GANs), a dominant paradigm for pixel-level image synthesis, have been widely adopted for lip-synchronized facial animation in prior research~\citep{b28,b2,b11,b12,b13,b14}. These methods typically adopt a conditional generation framework where (1) partially occluded facial inputs (masked lower-face regions) and raw audio streams are encoded as multimodal conditions, and (2) a generator network reconstructs lip-synchronized frames through adversarial training. Wav2Lip~\citep{b2} first introduce a lip-sync discriminator that optimizes audio-visual coherence. it advances the field through its adversarial framework that penalizes asynchronous lip motions via a pre-trained sync discriminator, enabling robust lip synchronization even for in-the-wild talking portraits. However, GAN-based methods often suffer from mode collapse~\citep{b22,bb5}, where diverse identity-specific details are lost, leading to a lack of personalized facial features and blurry or distorted outcomes.
\subsubsection{Two-stage Talking Face Generation.}
Current research paradigms in talking face generation have evolved beyond direct pixel-level lip synchronization. Several existing approaches ~\citep{bc3,b3,b4,b14,b15,b23} adopt a two-stage procedural framework: the first stage synthesizes audio-driven motion representations, typically encoded as either 3D facial morph coefficients or landmark displacement vectors; The second stage utilizes these representations within a 3D Morphable Model (3DMM) ~\citep{bb6, bc4} framework to reconstruct and animate facial geometries. For instance, MakeItTalk ~\citep{b3} employs a landmark displacement prediction module whose outputs drive a subsequent neural renderer for facial animation. Similarly, SadTalker~\citep{b4} first generates the 3D motion coefficients of lip motions, expressions, and head poses respectively, and these coefficients are modulated by a 3D-aware face render~\citep{bc4} for final video generation. While intermediate representation-based methods improve structural accuracy, they often struggle to retain the original speaker’s emotion and facial dynamics like eye-blinks and head poses. Consequently, methods relying on intermediary representations struggle with realistic identity retention, especially when the generated talking face is attached back to the full body. 


\subsection{Diffusion-based Talking Face Generation.}
Recent advancements in diffusion models have significantly advanced text-to-image (T2I) generation, surpassing generative adversarial networks (GANs) as the state-of-the-art approach for high-fidelity image synthesis~\cite{b31}. Inspired by the success of diffusion models in the T2I field, DiffusionVideoEditing ~\citep{b24} and Diff2lip ~\citep{b25} have adapted end-to-end denoising diffusion models for audio-driven video editing. These methods mask the lip region of the face frames and apply the denoising processes to inpaint the region in pixel space conditioned on audio spectral features. However, these methods take multiple denoising steps to inpaint face images, which makes it time-consuming and computation-intensive. 

To address this efficiency bottleneck, MuseTalk~\citep{b5} transitions the diffusion process from pixel space to latent space. The diffusion model employs a U-Net~\citep{b8} architecture with cross-attention mechanisms where audio features (encoded into a latent vector) are integrated into the denoising process via cross-attention modules within a Latent Diffusion Model (LDM). This latent space operation significantly reduces computational complexity by operating on lower-dimensional representations compared to raw pixel space. However, while latent-space diffusion improves efficiency, the LDM architecture inherently prioritizes high-level semantic fidelity over fine-grained texture preservation, leading to degraded resolution in synthesized lip contours and facial details.

\section{Method}

We propose UnAvgLip to achieve high identity-consistent face visually dubbing. The overview of UnAvgLip is shown in \cref{fig:framewrok}. Our framework is based on the open-source framework MuseTalk~\citep{b5}, where we leverage a pre-trained audio-conditional U-Net model as the neural backbone. The UnAvgLip consists of two components: Identity Perceiver and ID-CrossAttn module. The Identity Perceiver implements a Transformer-based architecture to project identity embeddings to a sequence of features, whose dimensions are the same as audio embeddings. The ID-CrossAttn module introduces the decoupled cross-attention strategy as IPAdapter~\citep{b35} to enhance the fine-grained control of identity details without weakening the correlations between visual representations and audio embeddings. We detail our proposed UnAvgLip in the following. 

\subsection{Conditional Latent Diffusion}
The backbone model can be seen as a conditional image-to-image LDM, which is used to inpaint the masked lip region conditioned on the audio feature. We leverage the power of U-Net's multi-scale fusion network with cross-attention mechanism to fuse audio and visual features across various scales and model the condition distributions of the form $p\left ( {z}  \mid a\right  )$.  As shown in \cref{fig:framewrok}, at time $t$, the VQ-VAE~\citep{b7,b34} encoder \( \varepsilon \) encodes the source image and the half-masked image \( I_{source}^{t}, I_{masked}^{t} \in \mathbb{R}^{H\times W \times3} \) 
into a latent low-dimensional representation \(z_{source}^{t}, z_{masked}^{t} \in \mathbb{R}^{h \times w \times c} \), which is concatenated to form the final latent representation \(z_t \in \mathbb{R}^{h \times w \times 2c} \). For the corresponding audio segment at time $t$ with window length of $T$, it is first re-sampled to 16,000 Hz and then converted into an 80-channel log magnitude Mel spectrogram \( mel_{t} \in \mathbb{R}^{T \times 80} \). We then use a pre-trained speech recognition model, Whisper~\citep{b9}, to encode Mel spectrogram \( mel_{t} \) into audio feature \( a_t \in \mathbb{R}^{T \times 384} \). The audio feature is fused to the middle layer \(i\) of U-Net by attention mechanism:

\begin{align}
\text{Attention}(\mathbf{Q},\mathbf{K},\mathbf{V}) = \text{Softmax}(\frac{\mathbf{Q}\mathbf{K}^{\top}}{\sqrt{d}})\mathbf{V} \\
      \mathbf{Q} = W^{i}_{Q} \cdot \varphi_{i}(z_t), \mathbf{K} = W^{i}_{K} \cdot a^t, \mathbf{V} = W^{i}_{V} \cdot a^t
\end{align}

where \(\varphi_{i}(z_t) \in \mathbb{R}^{N \times d_{\epsilon}^i} \) denotes a intermediate representation of the U-Net predicting \(\epsilon_{\theta}\) and  \( W^{i}_{Q} \in \mathbb{R}^{d \times d_{\epsilon}^i}, W^{i}_{K} \in \mathbb{R}^{d \times d_{\tau}}, W^{i}_{V} \in \mathbb{R}^{d \times d_{\tau}}\) are learnable matrices. Ultimately, the predicted visual latent feature \( \hat{z}_{t}^{w \times h \times c} \) is fed into a pre-trained VQ-VAE decoder to generate the bubbed image \(I_{gen} \in \mathbb{R}^{H\times W \times3}\).


\subsection{UnAvgLip}
We propose UnAvgLip to fuse additional identity embeddings extracted from a face detection model into the pre-trained UNet model. Our proposed UnAvgLip includes an Identity Perciever and ID-CrossAttn module. Identity Perciever projects the identity embeddings into a sequence of features that align with the dimension of audio features while ID-CrossAttn embeds identity information into the original U-Net model with decoupled cross-attention layers. We will detail the implementation in the following subsections.

\subsubsection{Identity Perceiver}

The architecture of Identity Perceiver is shown in \cref{fig:framewrok} (in light blue rectangle). It is inspired by the Perceiver ~\citep{b26} architecture , which effectively projects high-dimensional inputs into a fixed-dimensional output while retaining expressivity. The identity embeddings \( emb_{id} \in \mathbb{R}^{N \times 512} \) are first mapped to a sequence of identity tokens \(E_{id} \in \mathbb{R}^{N_q \times 384} \) through a linear projection (we set \( N_q = 4 \) in our experiments), aligning the dimension with the audio features. The model then alternates between applying a cross-attention module and a Transformer decoder layer, repeating this process L times (we set \( L= 3 \) in our experiments). Specifically, the high-dimensional input is projected through a low-dimensional attention bottleneck before being processed with the Transformer. The resulting representation is then used to query the input again. The identity query tokens act as a dynamic summary of the input, iteratively refining through layers of attention and Transformer decoders, selectively attending to and integrating relevant information from the face embeddings.




\subsubsection{ID-CrossAttn}
ID-CrossAttn integrates the identity query tokens into the original lip-syncing U-Net model by introducing decoupled cross-attention mechanism. In the backbone U-Net, it learns the correlation between visual features and audio features and models the conditional distribution domain by adopting condition-related cross-attention modules. To plug the global identity embeddings into the original U-Net model, we add a new cross-attention layer for each audio-visual cross-attention layer in the original U-Net model. Specifically, given the aligned identity embeddings \( E_{id} \in \mathbb{R}^{N \times 384} \) we obtained from Identity Perceiver, the decoupled ID-CrossAttn is defined as follows:

\begin{align}
\text{ID-CrossAttn}(\mathbf{Q},\mathbf{K_{if}},\mathbf{V_{id}}) = \text{Softmax}(\frac{\mathbf{Q}\mathbf{K_{id}}^{\top}}{\sqrt{d}})\mathbf{V_{id}} \\
      \mathbf{Q} = W^{i}_{Q} \cdot \varphi_{i}(z_t), \mathbf{K_{id}} = W^{id}_{K} \cdot E_{id}, \mathbf{V_{id}} = W^{id}_{V} \cdot E_{id}
\end{align}

Here, we use the same \(W_{Q}\) matrix as the original model and the new \(W^{id}_{K}\) and \(W^{id}_{V}\) are initialized from the original \(W_{K}\) and \(W_{V}\) to speed up the convergence. The \(W^{id}_{K}\) and \(W^{id}_{V}\) are shared for each cross-attention layer. Therefore, we only need to train the parameters of \(W^{id}_{K}\) and \(W^{id}_{V}\), which reduces a quite amount of compute resources. We simply add the output of identity cross-attention to the output of the audio condition cross-attention. The final cross-attention is defined as follows:

\begin{equation}
\mathbf{Z}^{new}=\text{Softmax}(\frac{\mathbf{Q}\mathbf{K}{\top}}{\sqrt{d}})\mathbf{V}+\lambda \cdot \text{Softmax}(\frac{\mathbf{Q}\mathbf{K_{id}}^{\top}}{\sqrt{d}})\mathbf{V_{id}}
\end{equation}
where $\lambda$ represents the scale factor, and the model becomes the original text-to-image diffusion model if $\lambda=0$. 

\subsection {Loss Function} 
In the training stage, we use four kinds of loss functions to train our UnAvgLip, including reconstruction loss, perception loss, and lip-sync loss.

\textbf{Reconstruction loss.} We crop the half-lower part of dubbed image \(I_o \in \mathbb{R}^{H \times W \times 3}\) into \(\hat{I}_o\in \mathbb{R}^{\frac{H}{2} \times \frac{W}{2} \times 3}\) and ground-truth image \(I_{gt}\in \mathbb{R}^{H \times W \times 3}\) into \(\hat{I }_{gt}\in \mathbb{R}^{\frac{H}{2} \times \frac{W}{2} \times 3}\). The reconstruction loss $L_{rec}$ uses L1 loss to measure the pixel-wise distance between dubbed image and ground-truth image. The reconstruction loss \(L_{rec}\) is written as:

\begin{equation}
\mathcal{L}_{rec} = \left \| \hat{I}_o - \hat{I}_{gt}  \right \|_1
\end{equation}

\textbf{Perception Loss.} Compared to reconstruction loss, perception loss~\citep{b36} focuses more on the perceptual quality of images, aligning better with human visual perception of image quality. The paired images {\(\hat{I}_o, \hat{I }_{gt}\)} are passed through a pretrained VGG-19 network ~\citep{b37} to obtain their feature representations. These feature representations are then used as inputs to the loss function. The perception loss is written as

\begin{equation}
\mathcal{L}_{p} = \left \| \mathcal{V}(I^{t}_{o}) - \mathcal{V}(I^{t}_{gt})  \right \|_2
\end{equation}
where $\mathcal{V}$ denotes the feature extractor of VGG19.

\textbf{Lip-sync loss.} We adopt a pretrained SyncNet to penalize inaccurate synchronization of lip movements. During the training, a window \(V\) of \(T_v\) consecutive dubbed face frames \( V \in \mathbb{R}^{N\times \frac{H}{2} \times 3\cdot T_{v}}  \) and a speech segment \(S \in \mathbb{R}^{N\times T_a \times D}\) are fed into the sync expert discriminator, where \(T_v\) and \(T_a\) are the video and audio timesteps respectively (we set \(T_v= 5\)  and \(T_a= 26\) in this study). The SyncNet generates the frames and audio embedding pairs {\(v,s\)}, then we use cosine-similarity with binary cross-entropy loss as Lip-Sync loss. The lip-sync loss is written as 

\begin{align}
    P_{sync} = \frac{v \cdot s}{\left \| v \right \|_2 \cdot \left \| s \right \|_2 }  \\
    \mathcal{L}_{sync} = \frac{1}{N} \sum_{i}^{N}-log(P_{sync})
\end{align}



\section{Experiment}

\subsection{Experiment Setup}


\subsubsection{Datasets}

We conduct our experiments on two publicly available datasets that are widely used for talking face generation: Lip Reading in the Wild (LRW) dataset~\citep{b32} and HDTF dataset~\citep{b18}. The LRW dataset consists of word pronunciation clips extracted from BBC videos, containing 500 different words spoken by hundreds of different speakers. Each video clip has a duration of approximately 1 second, comprising 29 frames. The HDTF dataset consists of over 400 in-the-wild videos, with resolutions of either 720P or 1080P. For training, we use the training set from the LRW dataset to train our UnAvgLip model. For evaluation, we randomly select 20 test videos from the test sets of both LRW and HDTF datasets to evaluate the model’s performance.

\subsubsection{Implementation Details.}
Our experiment is based on MuseTalk\footnote{\url{https://github.com/TMElyralab/MuseTalk}} framework. In data preprocessing stage, a facial landmark detection model is employed to get the face bounding box to crop the face regions of talking video frames and resize them into $256 \times 256$ resolution. We randomly select $N$ face images from cropped frames and use a pre-trained face recognition model~\citep{bb27} from the open-source InsightFace library to extract identity embeddings. For audio data, we use the pre-trained Whisper~\citep{b9} tiny model to convert melspectrogram into audio embeddings and split them into chunks corresponding to video frames. We add a new cross-attention layer wrapped in our ID-CrossAttn module for each cross-attention layer and initialize ID-CrossAttn module with parameters of the cross-attention layers in pre-trained UNet model. In the training stage, only Identity Perceiver and ID-CrossAttn modules with total 48M parameters are trained, which is relatively cheap to train compared with fine-tuning on the original model with 890M parameters. We conduct our experiment on 8 NVIDIA 3090 GPUs (24GB) for 10,000 steps with a batch size of 2 per GPU. We use the AdamW optimizer with a fixed learning rate of $10^{-5}$. To maintain guide-free inference, we drop the identity embeddings with a probability of 0.05.

\subsubsection{Evaluation metrics.}

We evaluate our method in three aspects: (1) Visual quality. We compute the metrics of Structural Similarity (SSIM) ~\citep{b21}  and Peak Signal to Noise Ratio (PSNR)~\citep{b16, b38}  (2) To evaluate the audio-visual synchronization, inspired from ~\citep{b2}, we compute the metrics of Lip Sync Error Distance (LSE-D) and Lip Sync Error Confidence (LSE-C) with open-source codebase\footnote{\url{https://github.com/joonson/syncnet_python}}. (3) To evaluate the identity retention capability, we calculate the cosine similarity (CSIM)~\citep{b5} between the identity embeddings of the source and generated images.  


\begin{table}[t]
\label{performance-metrics-htdf-mead}
\begin{center}
\resizebox{\textwidth}{!}{%
\begin{tabular}{lcccccccccc}
\midrule
\multicolumn{1}{c}{\bf } & \multicolumn{5}{c}{\bf HTDF} & \multicolumn{5}{c}{\bf LRW} \\
\cmidrule(lr){2-6} \cmidrule(lr){7-11}
& \multicolumn{1}{c}{SSIM$\uparrow$} 
& \multicolumn{1}{c}{PSNR$\uparrow$} 
& \multicolumn{1}{c}{LSE-C$\uparrow$} 
& \multicolumn{1}{c}{LSE-D$\downarrow$} 
& \multicolumn{1}{c}{CSIM$\uparrow$} 
& \multicolumn{1}{c}{SSIM$\uparrow$} 
& \multicolumn{1}{c}{PSNR$\uparrow$} 
& \multicolumn{1}{c}{LSE-C$\uparrow$} 
& \multicolumn{1}{c}{LSE-D$\downarrow$} 
& \multicolumn{1}{c}{CSIM$\uparrow$} 
\\
\hline \\
Wav2Lip ~\citep{b2} & 0.8287  &  25.669 & \textbf{8.463}  & \textbf{8.798}  & 0.8397 
& 0.8597  & 27.2857 & \underline{7.5674}   & \underline{7.5853}  & 0.8419 \\
VideoRetalking~\citep{b23} & 0.8011  & 23.5394 & 5.3965  &  10.56 & 0.8145
& 0.8057  & 24.6428 & 6.3798 & 9.2929 & 0.8296 \\

TalkLip ~\citep{b38} & 0.8421 & 26.9654 & \underline{7.46}  & \underline{9.11} & 0.8285
& \bf{0.9136}  & \bf{30.0966} & \bf{8.7989} & \bf{6.74} & 0.8420 \\

SadTalker ~\citep{b4} &0.6410  &  22.0709 & 3.5699  & 11.8685 & 0.7577
& 0.7449   & 20.8 & 4.8264 & 10.8404 & 0.7341 \\

MuseTalk ~\citep{b5}  &  \underline{0.8501 }& \underline{28.5923} &   6.3533 & 8.1188   & \underline{0.842}  
& 0.8775  & 29.7966 &  6.318 & 8.2351  & \underline{0.8535}  \\

\midrule
Ours & \textbf{0.8753} & \textbf{29.6098} & 6.9039   & 7.6992 & \textbf{0.8867}
& \underline{0.8971}  & \underline{29.731} & 7.031 & 8.1079 & \bf{0.9076} \\
\bottomrule
\end{tabular}%

}
\end{center}
\caption{Quantitative evaluations with the state-of-the-art methods on visual bubbling task on HDTF and LRW test set. Bold and underline correspond to the optimal and sub-optimal values, respectively.}
\label{table:quantative_results}
\end{table}

\begin{figure}[htpb]
    \centering
    \includegraphics[width=\textwidth]{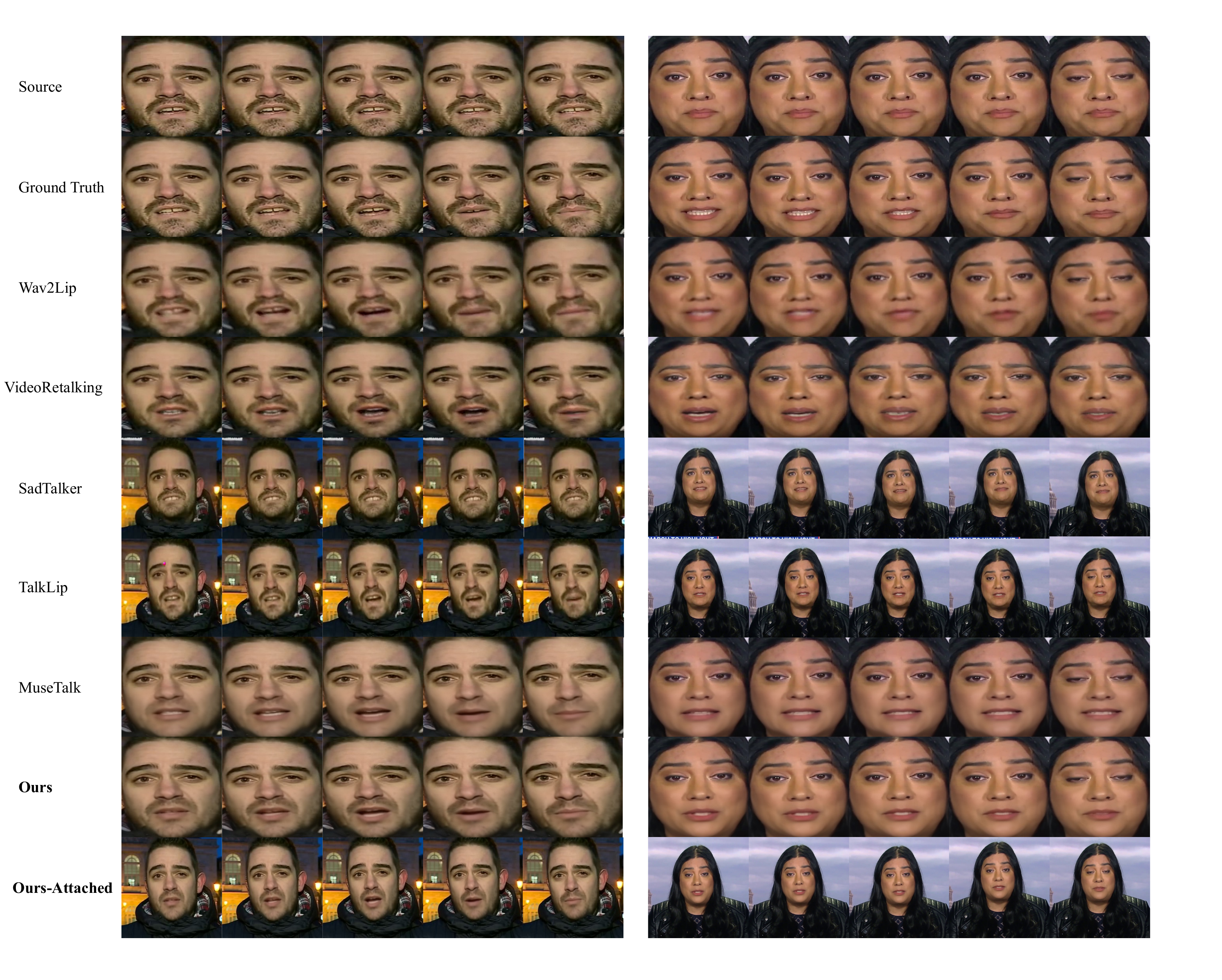}
    \caption{
    Qualitative comparisons with SOTA lip-sync methods. We evaluate our method on two test cases from the dataset. The first row presents the source frames to be edited, while the second row shows the target frames, representing the desired lip movements. Rows 3 to 6 display the results generated by various SOTA lip-sync methods, and the last row presents the outputs produced by our method. (All photorealistic portrait images in this paper are sourced from licensed models.)
    }
    \label{fig:qualitive_results}
\end{figure}


\subsection{Comparisions}

\subsubsection{Compared Baselines.}

we compare our method with five state-of-the-art (SOTA) open-source lip-syncing frameworks: Wav2Lip~\citep{b2}, VideoRetalking~\citep{b23}, SadTalker~\citep{b4}, TalkLip~\citep{b38}, and MuseTalk~\citep{b5}. 

Wav2Lip is a pioneering framework in the domain of arbitrary-identity lip-audio synchronized face generation using adversarial training. It introduces a lip-sync expert to train the generator and serves as a key benchmark for evaluating lip synchronization performance on dubbed videos. VideoRetalking generates talking face videos through a three-step process: it first neutralizes the original facial expression, then generates a talking head video via a lip-sync network, and finally applies StyleGAN to enhance the visual quality. SadTalker generates one-shot talking faces in a two-stage process: 1) it uses a 3D morphable model (3DMM) to extract intermediate representations and generates 3D coefficients driven by audio, facial expressions, and head pose; 2) it applies a 3D-aware face renderer to animate and render the reference face into a realistic talking head video. TalkLip leverages a fine-tuned lip-reading model from AV-Hubert during training to supervise the generator for dubbed frames and introduces contrastive learning between visual representations and audio embeddings to improve lip-speech synchronization. MuseTalk, on the other hand, is an image-to-image inpainting network based on Latent Diffusion Models (LDM), where the backbone UNet performs multi-scale modality alignment between visual representations and audio features, resulting in lip-synced talking face generation. 


\subsubsection{Quantitative comparisons.}


As observed in the quantitative results \cref{table:quantative_results}, Wav2Lip and TalkLip demonstrate strong and stable performance in lip synchronization. On the HDTF dataset, UnAvgLip and MuseTalk achieve superior visual quality and identity consistency compared to other models, whereas TalkLip performs better on the LRW dataset in terms of visual quality but struggles with high identity consistency. SadTalker, constrained by its one-shot generation approach, exhibits lower scores across all metrics. Notably, our UnAvgLip consistently achieves the highest CSIM scores across both datasets, providing strong evidence of its effectiveness in enhancing identity consistency while maintaining competitive performance in lip synchronization and visual quality.

\subsubsection{Qualitative comparisons.} \label{Section IV}
 To qualitatively compare different methods, we select two videos from 
 generated talking faces using different methods for two videos selected from the test set of two datasets respectively. The qualitative results are shown in \cref{fig:qualitive_results}. The source frames and the ground-truth frames are placed in the first and second rows following the dubbed face frames of different methods. While the Wav2Lip performs well in term of lip synchronization, it generates blurry results in higher resolution 1080P video.  VideoRetalking generates talking face frames with higher clarity. However, it could generate wrong lip shape or unnatural lip and teeth which diminishes the identity consistency and the lifelike quality of the dubbed video. SadTalker generates photo-realistic talking video but can lead to unnatural facial expression and head pose and limits the motion range to the upper part of the neck. While TalkLip excels at lip synchronization, a bounding box can be seen in the dubbed video. Although MuseTalk shows relatively better performance on identity retention, an “averaging lip” phenomenon can still be observed. As contrast, our  UnAvgLip generates more precise lip shape and more fine grained facial details, demonstrating superior performance on identity retention capability and generation visual quality.

\subsection{Ablation studies}

We conduct ablation studies to validate each module in our UnAvgLip framework. Specifically, we remove the ID-CrossAttn module and replace the Identity Perceiver with a linear projection model. Additionally, we train both of these modules without SyncNet to demonstrate the necessity of the synchronization loss. Furthermore, we perform experiments with different combinations of the number of reference images and weight factors to find out the optimal configuration.

\begin{figure}
    \centering
    \includegraphics[width=0.8\linewidth]{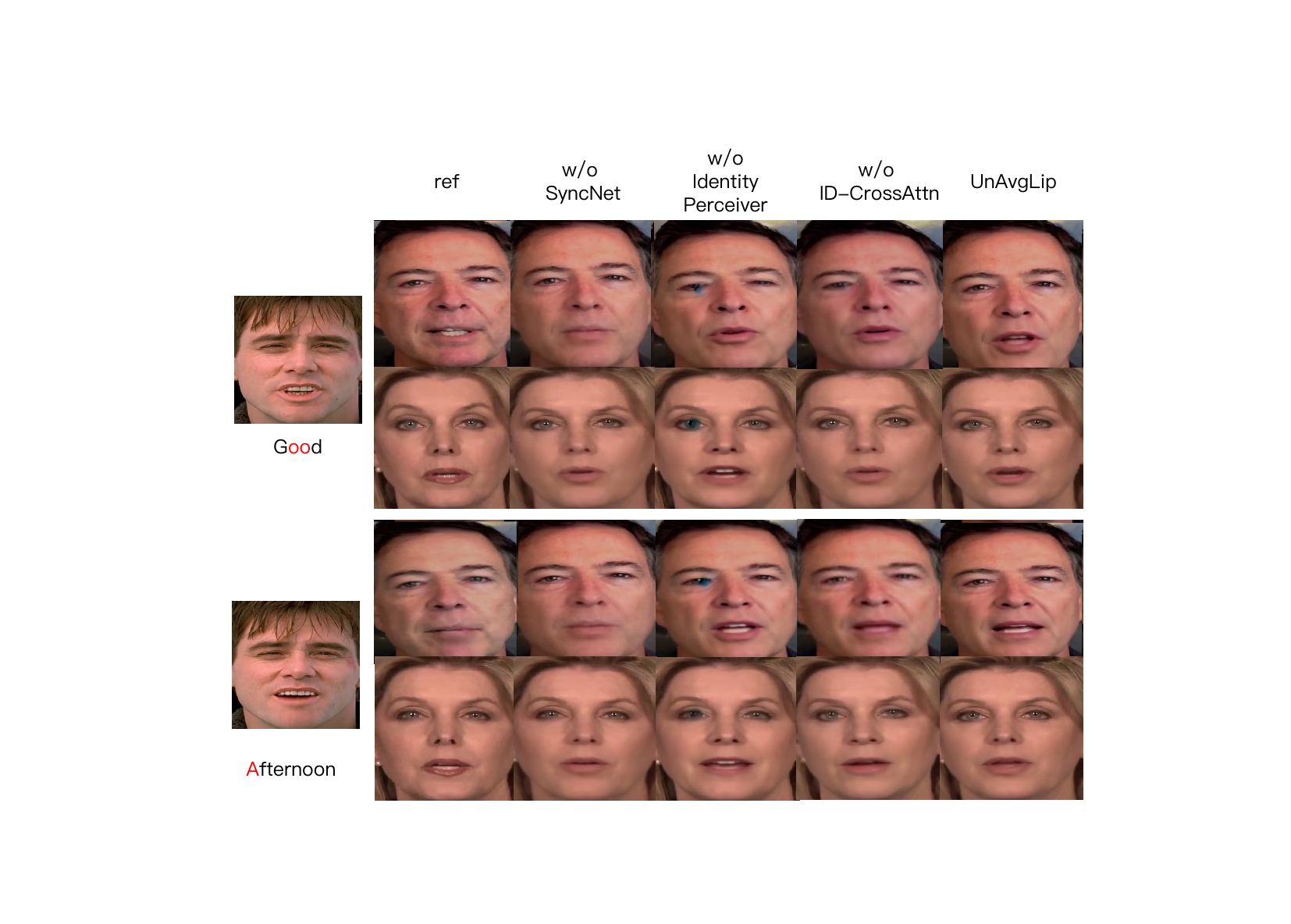}
    \caption{Qualitative results of ablation study.}
    \label{fig:ablation}
\end{figure}

\begin{table}[ht]
    \centering
     \resizebox{1.0\columnwidth}{!}{
    \begin{tabular}{lccc}
    \toprule
         & CSIM $\uparrow$ & LSE-D $\downarrow$ & LSE-C $\uparrow$ \\
         \midrule
        \textbf{UnAvgLip} & 0.8754 & \textbf{7.6992} & \textbf{6.9039}  \\
        - w/o ID-CrossAttn & 0.8314 & 8.1188 & 6.3533  \\
        - w/o Identity Perceiver & 0.8221 & 8.1079 & 7.5674  \\
        - w/o SyncNet & \textbf{0.8798} & 11.7838 & 1.6537 \\
        \bottomrule
    \end{tabular}
    }
    \caption{The ablation studies on HDTF dataset.}
    \label{table:ablation}
    \vspace{-1mm} 
\end{table}

\subsubsection{Identity Perceiver or Linear Projection Only?}

To assess the effectiveness of the proposed Identity Perceiver, we replace it with a simpler Linear Projection model consisting of a single linear layer. The linear projection module is trained alongside the Adapter modules for the same number of training steps (10,000).

As shown in \cref{table:ablation}, using only a single linear layer to project identity embeddings into the same dimension as the audio features leads to a noticeable degradation in visual quality, even compared to the baseline model. In some cases, it introduces ghosting artifacts in the unmasked regions (see \cref{fig:ablation}). This suggests that a simple linear transformation is insufficient for preserving detailed identity information.

One possible explanation is that the identity embeddings have a significantly higher dimensionality than the audio features. A single linear layer fails to extract and query the most relevant identity information from original identity embeddings effectively, resulting in a loss of speaker-specific facial details. In contrast, our proposed Identity Perceiver, which leverages the attention mechanism, is able to dynamically capture and inject the most relevant identity features as additional conditioning for the U-Net model, thereby enhancing identity consistency while maintaining high-quality lip synchronization.

\subsubsection{A Lip Sync Expert is needed.}
We trained our Identity Adapter without lip-sync loss. After it is trained, we found it can't generate accurate lip movements synchronizing with the target audio (see \cref{fig:ablation}) and it just generates the frames almost same with the reference frames, showing no lip sync ability at all. After adding lip sync loss to the total loss, the Identity Adapter is able to maintain the lip synchronization ability. We assume that the supervision of SyncNet acts like a regularization which prevents the model from over-fitting.

\begin{figure}[ht]
    \centering
    \begin{subfigure}[b]{0.45\textwidth}
        \includegraphics[width=\textwidth]{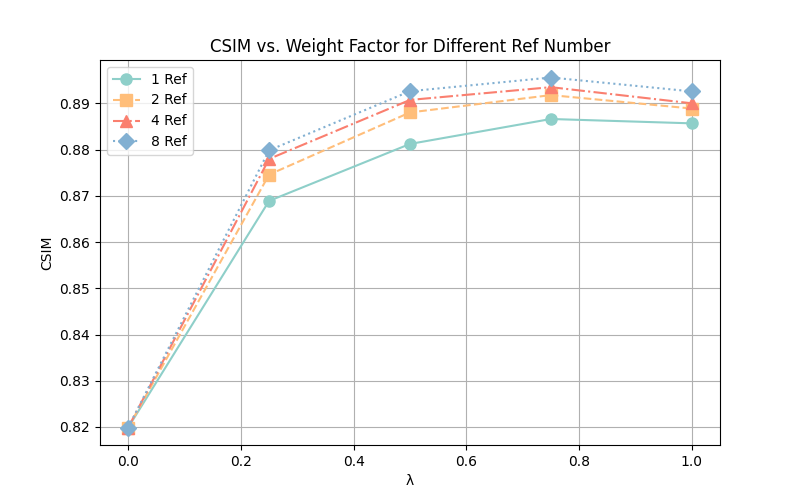}
        \caption{Quantitative Effect of Weight Factor and Reference Number on Identity Similarity.}
        \label{fig:csim_vs_scale_ref_number}
    \end{subfigure}
    \hfill
    \begin{subfigure}[b]{0.45\textwidth}
        \includegraphics[width=\textwidth]{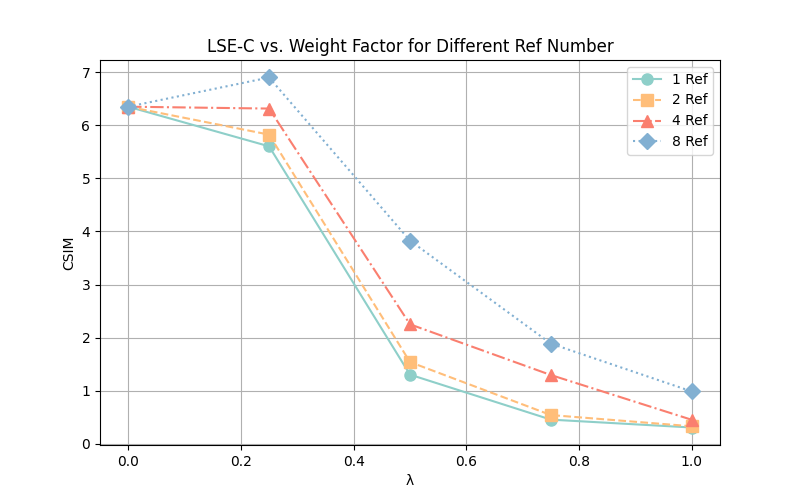}
        \caption{Quantitative Effect of Weight Factor and Reference Number on Lip Synchronization.}
        \label{fig:lsc_vs_scale_ref_number}
    \end{subfigure}
    \caption{Effect of Weight Factor and Number of Reference Faces.}
    \label{fig:subfigures}
\end{figure}

\begin{figure}[thbp]
  \centering
    \includegraphics[width=\linewidth]{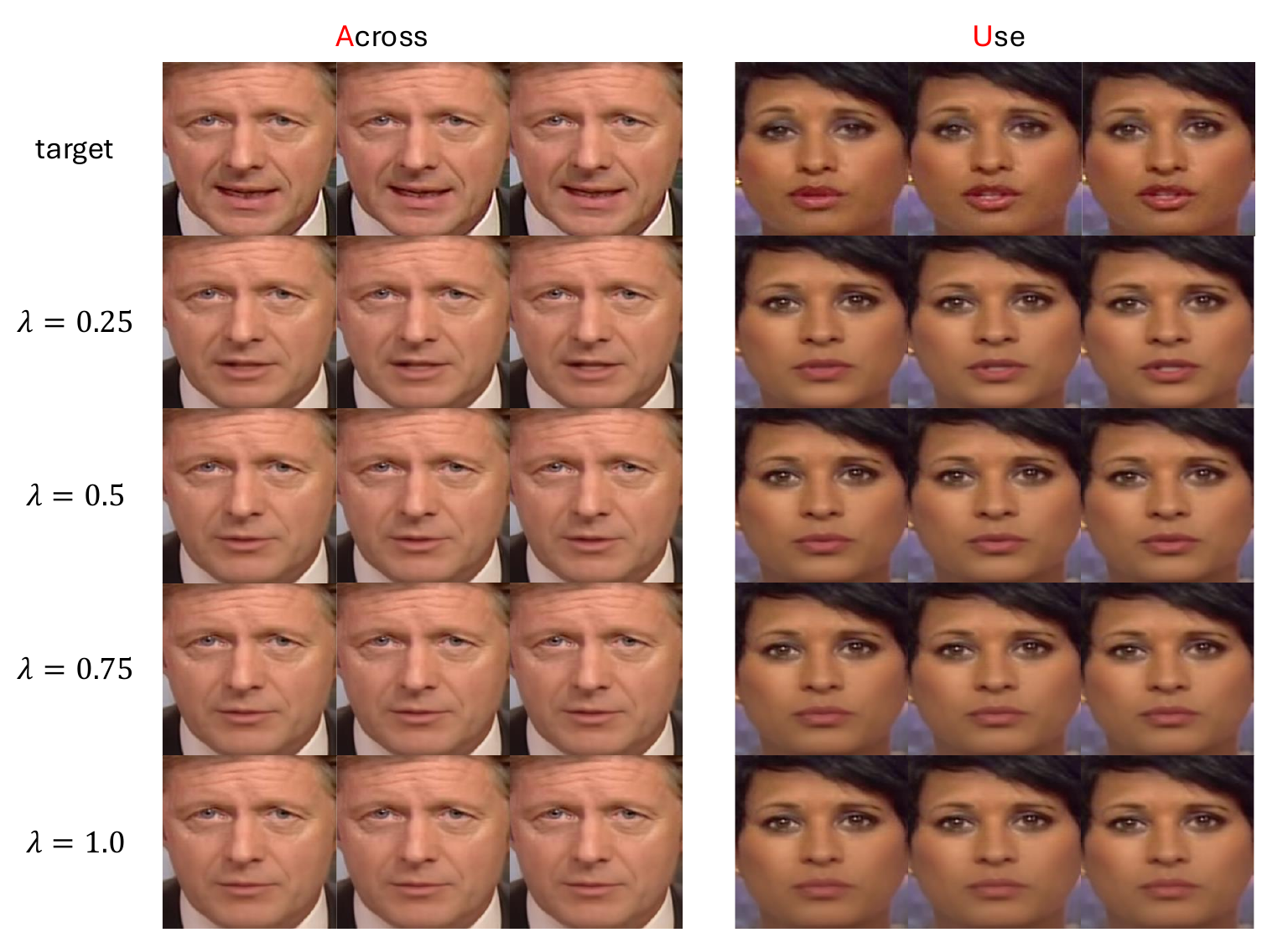}
  
  \caption{\textbf{Effect of Weight Factor on Lip Motion Editability.} As the weight factor increases, the lip motion becomes less synchronized with the audio.}
  \label{fig:effect_of_weight_factor}
\end{figure}

\subsubsection{Effect of Weight Factor and Reference Face Number}
Although ID-CrossAtten plugs identity embeddings into the backbone U-Net model for making the model attend to global identity information, it simultaneously distracts the original cross-attention mechanism between the visual representations and audio features, resulting in negative effect on lip synchronization. In this section, we adjust the impact of the newly introduced ID-CrossAtten on the overall cross-attention mechanism by tuning the weight factor. This allows us to balance the model's ability to maintain identity consistency and lip-syncing capability. As shown in the \cref{fig:effect_of_weight_factor} , as the weight factor \(\lambda\) increases (from 0 to 1.0), the CSIM score between the generated facial video frames and the original video frames increases. However, the LSE-C score, which represents lip-syncing ability, drops significantly. Notably, when the \(\lambda\) factor exceeds 0.5, the LSE-C score experiences a sharp decline. As illustrated in Figure 5, when \(\lambda\) is greater than 0.5, the generated facial lips remain almost entirely closed, with only slight lip movements. When \(\lambda\) is set to 0.25, the model's LSE-C score is comparable to, or even slightly better than, the original model. These results demonstrate that, at \(\lambda = 0.25\), ID-CrossAtten enhances identity consistency while maintaining the lip-syncing capability of the original model.

\begin{figure}[thbp]
  \centering
    \includegraphics[width=\linewidth]{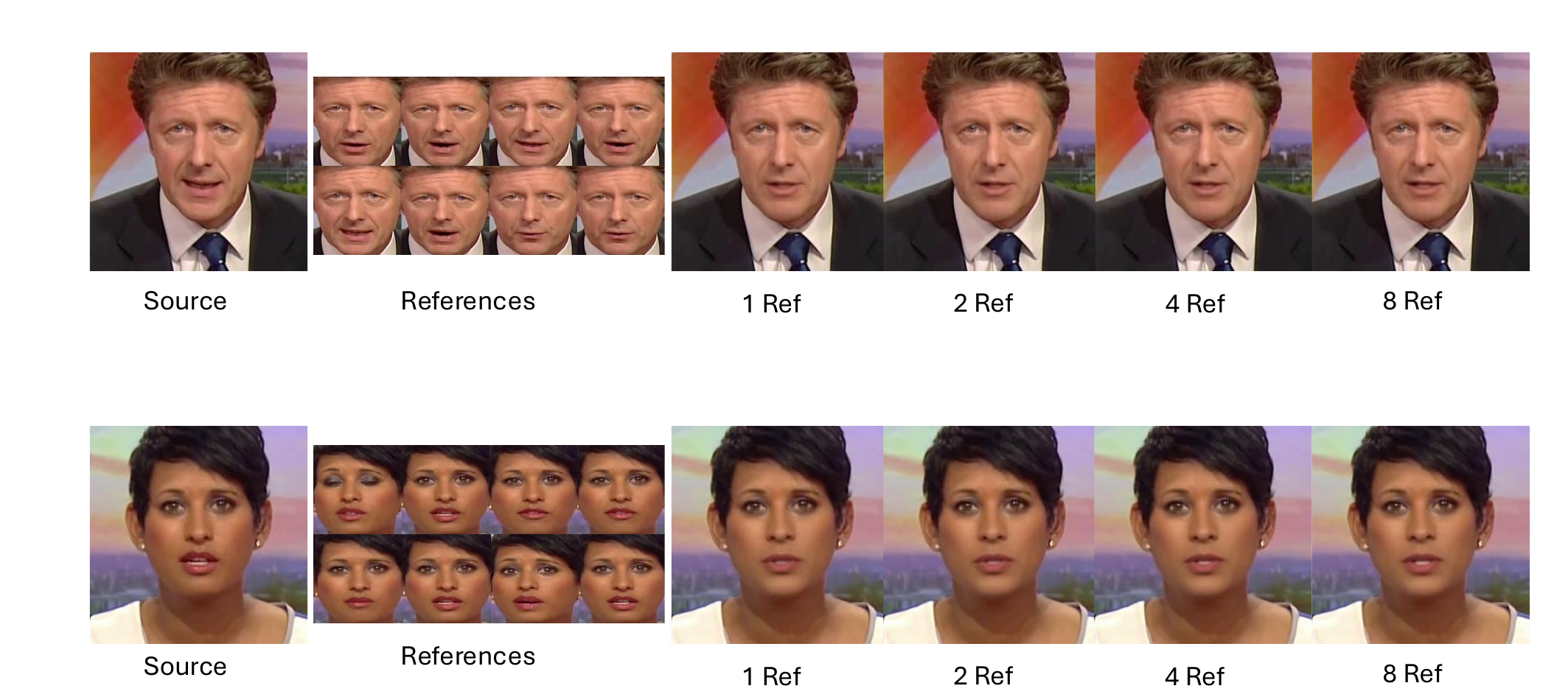}
  
  \caption{\textbf{Effect of Number of Reference Images.} We select N (N=1, 4, 8) frontal-view facial images as references and use the mean of identity embeddings as the global identity reference.}
  \label{fig:effect_of_reference_number}
\end{figure}

Additionally, we investigate the effect of the number of reference face images on the model's ability to preserve identity consistency. We randomly select \(N\) reference face images from the input video frames and extract identity embeddings from them. The average of these embeddings is then used as the global ID information to guide the generation of the talking face. As shown in \cref{fig:effect_of_reference_number}, using multiple reference images improves the identity consistency of the generated faces. Ultimately, we adopt \(\lambda = 0.25\) and \(N = 4\) as the default settings for our experiments.

\section{Conclusion}
In this paper, we introduce UnAvgLip, a novel approach that effectively addresses the ``lip averaging'' issue in talking face generation through a trainable, plug-and-play ID-CrossAttn module. By explicitly integrating identity embeddings into the synthesis process, our method enables high-fidelity, personalized talking-face generation while maintaining identity consistency.

One of the key advantages of UnAvgLip is its ability to capture and preserve speaker-specific facial features using only a few reference images—or even a single image—ensuring that the generated results retain the individual's unique identity throughout the audio-driven synthesis. Moreover, our framework achieves high visual quality while maintaining a relatively low computational cost, making it both practical and efficient for real-world applications.

Overall, UnAvgLip stands out for its efficiency, flexibility, and superior synthesis quality, offering a promising new direction for multimodal generation, identity preservation, and controllable face synthesis. We believe our work provides valuable insights that will inspire further advancements in visual dubbing, AI-driven avatars, and identity-aware generative models.

\section{Limitation}

While the incorporation of additional cross-attention for global identity information proves effective in enhancing identity retention, several challenges remain. The ID embedding is extracted from a pre-trained face recognition model, which contains rich semantic information such as age, gender, and eye color. However, this identity embedding may include coupled facial attributes irrelevant to the visual dubbing task. Given that visual dubbing primarily focuses on the lip shape, beard, and teeth in the lower half of the face, a more customized face encoder specifically designed for this task could better align with the needs of visual dubbing. Another key limitation lies in the sensitivity of the results to the choice of reference facial images. In our experiments, we used only frontal-view images to train and evaluate the model, which may restrict its performance when dealing with different poses or angles. This suggests the need for a more robust approach that can generalize across various viewpoints. Furthermore, synthesizing speech-coherent facial expressions remains a significant challenge. Achieving seamless integration of facial movements with speech is critical to making the generated video both expressive and natural. In future work, we plan to explore methods for better controlling facial expressions to ensure that the visual dubbing output appears more dynamic and realistic.







\end{document}